# SparCA: Sparse Compressed Agglomeration for Feature Extraction and Dimensionality Reduction

Leland Barnard    Farwa Ali    Hugo Botha    David T. Jones

Neurology Artificial Intelligence Program

Mayo Clinic

**Abstract**

The most effective dimensionality reduction procedures produce interpretable features from the raw input space while also providing good performance for downstream supervised learning tasks. For many methods, this requires optimizing one or more hyperparameters for a specific task, which can limit generalizability. In this study we propose *sparse compressed agglomeration* (SparCA), a novel dimensionality reduction procedure that involves a multistep hierarchical feature grouping, compression, and feature selection process. We demonstrate the characteristics and performance of the SparCA method across heterogenous synthetic and real-world datasets, including images, natural language, and single cell gene expression data. Our results show that SparCA is applicable to a wide range of data types, produces highly interpretable features, and shows compelling performance on downstream supervised learning tasks without the need for hyperparameter tuning.

## 1 Introduction

Dimensionality reduction (DR) is an important approach to data analysis and statistical learning tasks and is utilized across every discipline that deals with high dimensional data. DR can eliminate collinearities and redundancies from data and counteract the curse of dimensionality, which can improve performance on supervised learning tasks. Without a supervised objective, DR can reveal latent patterns and structure in data. At a high level, most DR strategies involve a combination of two goals: feature compression of the raw high dimensional signal into a lower dimensional latent representation, and feature subset selection from among the starting feature set. Both strategies produce a lower dimensional representation of the input data but have different advantages and motivations. Feature compression can extract new features that are, in some cases, highly revealing about the underlying nature of the data. For example, independent component analysis can be used to separate a signal with many simultaneous inputs into separate constituent signals, such as in the classical application of separating a single recording of many speakers into separate signals for each speaker [1, 2]. However, in other circumstances feature compression methods extract features that are difficult to interpret because each extracted feature is a function of the entire input feature space. Even in the case of principal component analysis (PCA) where that function is simply a linear combination of the input features, the resulting features are generally not interpretable except in certain applications such as image analysis where the components can themselves be represented as images. Feature subset selection on the other hand produces a highly interpretable solution, as the result is reduced feature space, but the features themselves are untransformed from their input form. Pure feature subset selection is most often applied in a supervised learning setting through iterative methods or a regularization scheme that penalizes the L1-norm of the feature space [3, 4]. In unsupervised applications, feature selection is usually achieved by including a regularization penalty to the objective of a feature compression algorithm, as in the formulation of sparse PCA and related methods [5, 6]. In the conventional formulation of sparse PCA, each component is a linear combination of a sparse subset of the input features, which makes each component more easily interpretable. However, in general each component will use a different set of features, thus in aggregate the set of features may not be reduced. Recent evolutions of the sparse PCA method seek to address this problem by the additional constraint that each component use a common subset of features, thereby achieving feature compression and subset selection simultaneously [7]. A second drawback of sparse PCA relative to PCA is that it requires forming the feature space covariance matrix, whereas PCA can be computed with either subject space or feature space covariance. This makes sparse PCA computationally challenging for very high dimensional applications, even if the number of samples is relatively few.

Nearly all DR approaches involve one or more hyperparameters, and one of the challenges that must be addressed when applying a DR method is optimizing these parameters. For a supervised learning task, these hyperparameters can be optimized via cross validation, which is conceptually straightforward but increases the complexity of the hyperparameter space that must be searched. Without a supervised learning objective, choosing hyperparameters is more difficult. Even for PCA, perhaps the most well-known and understood approach to DR, rigorously choosing the optimal number of components is a difficult theoretical task with many different approaches [8]. While PCA has only one hyperparameter, other more complex methods such as UMAP [9] can have several, and choosing their values without a supervised learning objective is generally a manual process based on practitioner judgement.

In this study, we propose a novel approach to DR that performs both feature compression and selection to produce an interpretable reduced feature set that is sparse in the input feature space and has only two intuitive hyperparameters, one which may be confidently left with a default value and another which can be chosen by simple heuristic. Our proposed approach, called sparse compressed agglomeration (SparCA), is a judicious combination of three well-understood techniques: agglomerative clustering of features by Ward's method [10], compression of each cluster of features via PCA, and orthogonal matching pursuit (OMP) [11] to produce a sparse representation of each cluster component. Details are provided in the following section. We then demonstrate the efficacy of this approach on benchmark high dimensional datasets from three applications: image analysis, natural language processing, and single cell gene expression.

## 2 Methods

### 2.1 Basic description and formulation

The goal of SparCA is to perform DR by first grouping features into interpretable clusters or topics, compressing the features in each topic down into a relatively few principal components, and constructing a sparse representation of these components from the constituent features of each cluster. This procedure is illustrated in Fig. 1.

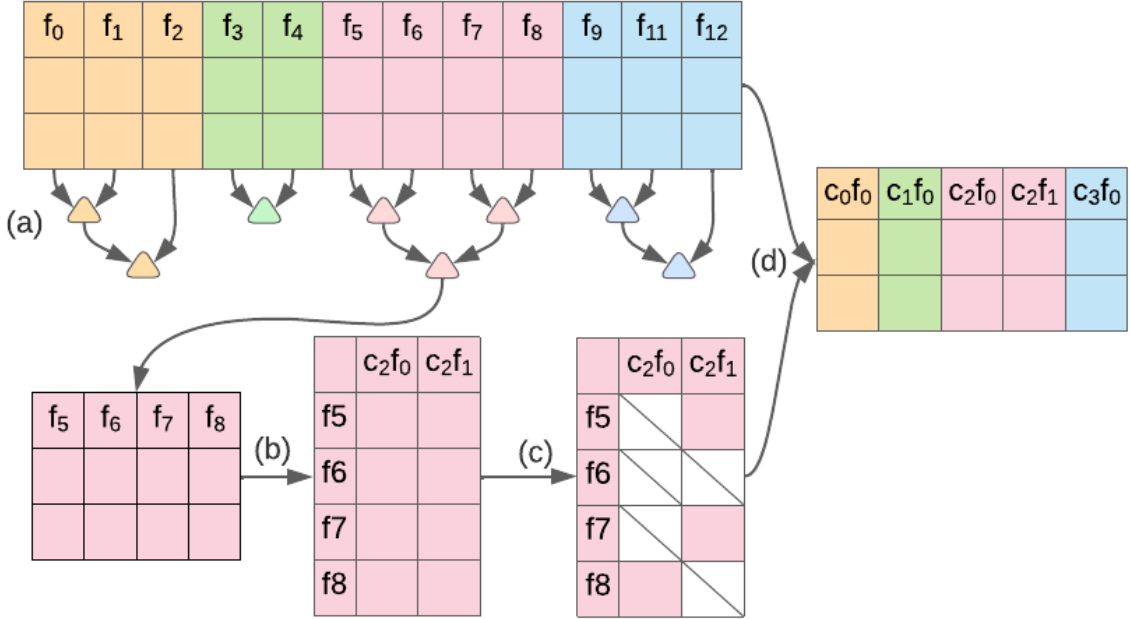

Figure 1: SparCA workflow. Features (represented by $f_0$-$f_{12}$) are clustered agglomeratively until the specified number of clusters is reached (a). Each cluster is compressed into a reduced set of features (represented for cluster 2 above as $c_2f_0$, $c_2f_1$) via PCA, where the number of components is determined via Horn's method (b). Orthogonal matching pursuit is used to resolve a sparse representation of each component, using only the minimum required input features to recover a specified fraction of the variance of the component loadings (c). The resulting components from each cluster are stacked together and multiplied with the input matrix, producing the lower dimensional feature set (d).

This approach differs from sparse PCA in that each feature is a sparse combination of features drawn from disjoint feature clusters, rather than from the full feature space. The first step in the procedure is performed using agglomerative clustering with Ward's minimum variance criterion (a), which operates pair-wise across the feature set and so can scale to very high dimensional datasets without reaching memory limitations. The number of clusters is the first of two hyperparameters that must be chosen for the SparCA procedure. Once the features have been grouped into a particular number of clusters, the next step in the procedure is performed using PCA to create a compressed representation of the features in each group (b). The number of components kept in each group is chosen automatically via Horn's method [12]. After each group of features has been compressed to a small number of principal components, OMP is used to construct maximally sparse approximations that recover a specified minimum variance of these components (c). This variance threshold is the second SparCA hyperparameter, although in practice it can be set to a default value between 0.9 and 0.99. In all analyses presented here it is set to a fixed value of 0.95. Finally, the sparse compressed components are applied to the input to produce the reduced feature space (d). Theoretical details of each element of the procedure are provided in the appendix. These steps are summarized in algorithm 1.

---

Algorithm 1: SparCA dimensionality reduction procedure

Inputs:
1. Feature matrix $X$ with dimensions $n$ samples by $m$ real valued features with zero mean and unit variance.
2. Number of feature clusters $N_c$.
3. Minimum recovered variance threshold $f$.

Output:
1. Linear feature transformation matrix $\Gamma$ with dimensions $m$ features by $p < m$ reduced features.
2. Compressed feature matrix $\tilde{X}$ with dimensions $n$ samples by $p$ reduced features.

```
1: input X, N_c, f
2:     C ← agglomerative_cluster(X, N_c)
3:     for i, c_i in C:              //cluster i with feature set c_i
4:         x_i ← X[:, feature ∈ c_i]
5:         h ← horn_analysis(x_i)   //determine number of components
6:         x̃_i ← pca(x_i, h)         //compress cluster to ℝ^h via pca
7:         γ_i ← omp(x_i, x̃_i, f)    //fit sparse components via OMP
8:         Γ ← column_concat(Γ, γ_i) //stack cluster components
9:     X̃ ← matrix_multiply(X, Γ)    //transform input matrix
10: return X̃, Γ
```

**2.2 Hyperparameter selection**

SparCA has two hyperparameters that must be chosen: the number of clusters into which to group the input feature set and the variance threshold for the OMP procedure. The latter controls the sparsity of the derived components from each cluster and can generally be set to a value between 0.9 and 0.99 and not tuned further. Hyperparameter selection therefore primarily entails selecting the number of clusters. SparCA differs from many other DR methods in that while the number of feature clusters is a chosen parameter, the number of compressed features that arise from those clusters is not and instead arises from the covariance structure of the data via Horn's method. The exact relationship between number of clusters and the resulting number of features is of course dataset-dependent, but our experiments show empirically that this relationship is in general a variation on a common shape: first, the number of features grows very rapidly with increasing number of clusters. As the number of clusters increases, this growth rate slows and the curve becomes nearly flat, such that for a range of cluster numbers, the number of features is approximately constant. Finally, beyond a certain threshold the number of features begins to increase again in linear proportion to the number of clusters, such that each new cluster adds exactly one new feature. This cluster-feature (c-f) curve is illustrated schematically in Fig 2, and real examples from the benchmark datasets analyzed in this study can be found in the following section. A good choice for the number of clusters for a given dataset can be made using this characteristic c-f curve according to the following simple heuristic: the number of clusters chosen should be the minimum number required to reach the constant plateau regime of the c-f curve.

**2.3 Datasets used and experimental details**

The behavior of the SparCA method was evaluated experimentally on three disparate high dimensional data types: images, free text, and single cell gene expression. For imaging, we chose MNIST [13], a well-known and commonly used dataset consisting of 70,000 images of handwritten digits. For free text, we used the IMDb movie review corpus [14], a dataset consisting of 50,000 movie reviews and corresponding ratings. For single cell gene expression, we analyzed brain tumor cell microarray data from the Curated Microarray Database (CuMiDa) [15], consisting of approximately 33,000 gene expression measurements from 130 cells across five tissue classes: ependymoma, glioblastoma, medulloblastoma, pilocytic astrocytoma, and normal. For each case, the analysis consisted of first learning a low dimensional representation of the data and then training a one-vs-rest multiclass logistic regression model with L1 regularization to perform a downstream classification task using the reduced feature space: digit recognition in the MNIST dataset, sentiment analysis in the IMBb review dataset, and cell type classification in the microarray dataset. To accomplish this, each dataset was first split into embedding and classification sets. The embedding set was used to train the DR model, and the classification set was further subdivided and used to train, optimize, and evaluate the classifiers. The resulting models were evaluated based on classification performance and interpretability. For each dataset, this analysis was performed using both SparCA as well as standard PCA to establish a baseline for comparison. For SparCA, the number of clusters was chosen based on the characteristic c-f curve as described in section 2.2. Rather than make this choice manually, for each dataset the c-f curve was numerically differentiated, and the number of clusters was chosen from the location of the minimum of the resulting first derivative curve. For PCA, Horn's method was used to establish the number of principal components. Thus, for both SparCA and PCA, the DR model hyperparameters were chosen prior to the supervised learning step and were not in any way optimized for classification performance. Additional analysis details specific to each dataset are provided below.

### 2.3.1 MNIST

The 70,000 images in the MNIST dataset were split into embedding, training, and test sets of 10,000, 59,000, and 1,000 images, respectively. Each 28x28 image was flattened into a 1D 784 vector and standard scaled using the mean and standard deviation of the embedding set. The embedding set was then used to train the SparCA and DR models, which were then applied to the training and test sets. The training set was used to train the classifiers and optimize the L1 shrinkage parameter via fivefold cross validation, and performance was then evaluated in the test set.

### 2.3.2 IMDb

The IMDb dataset consists of 50,000 free text samples and corresponding positive or negative sentiment labels. These text samples were first minimally processed to remove html and special characters, as well as stop words. The remaining text was then stemmed to create the final vocabulary. The processed documents were split into 10,000 embedding, 39,000 training, and 1,000 test examples. Features were extracted from the processed text via term frequency-inverse document frequency (TFIDF), with a term frequency threshold of 0.02 and n-gram range of 1 to 3. This resulted in a feature vector of length 978 for each document. As with the MNIST dataset, the embedding set was used to normalize the feature vectors and train the DR models, the training set was used to train the classifiers and optimize the L1 shrinkage parameter via fivefold cross validation, and performance was then evaluated in the test set.

### 2.3.3 CuMiDa brain tumor microarray data

The brain tumor microarray data was obtained from CuMiDa, an open source curated database of cancer microarray data sourced from the Gene Expression Omnibus [Edgar 2002, Barrett 2013]. This particular dataset was acquired using Affymetrix Human Genome U133 Plus 2.0 Array which includes markers for approximately 33,000 genes, and includes data from five different cell types. Additional details regarding experimental acquisition of the gene expression data are available from [16]. The 130 cells in the dataset were split evenly into embedding and classification sets of 65 cells each, and the measured expression values were standard scaled by the mean and standard deviation of the embedding set. The PCA and SparCA models were then fit using the embedding set and applied to the classification set. Due to the relatively few examples present in this dataset, the L1 shrinkage parameter was fixed at a constant value of 0.1 and classification performance was evaluated by fivefold cross validation in the classification set.

## 3 Results and discussion
### 3.1 Characteristic cluster-feature curves

Prior to developing classification models for each task, the number of clusters parameter must be selected for each dataset. This was accomplished in each case by following the procedure described in section 2.2. Fig. 2 shows the characteristic c-f curves for each dataset. For the MNIST and IMDb datasets, this curve was generated for a range of subsamplings of the embedding set, but was only generated once for the CuMiDa dataset due to the limited sample size.

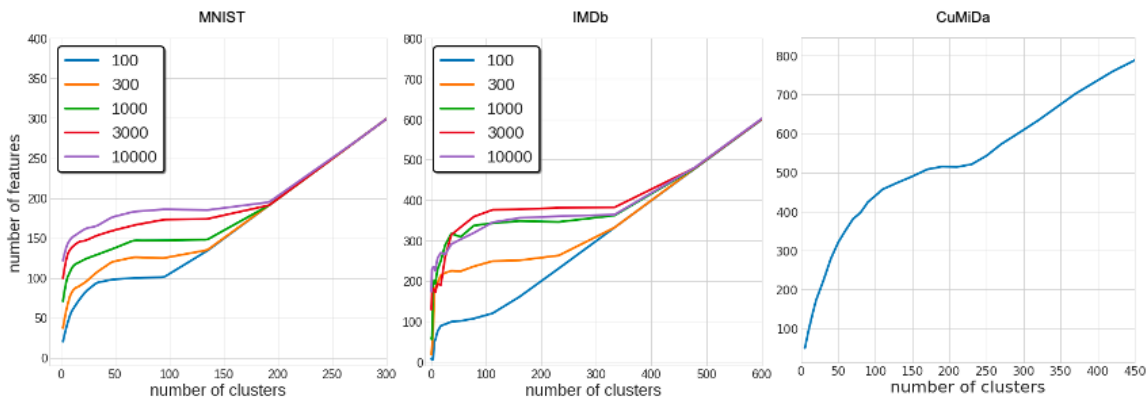

Figure 2: Characteristic cluster-feature curves for the three datasets studied

The curves are qualitatively similar across all three datasets, and within MNIST and IMDb the curves result in a consistent choice of cluster number across different subsample sizes. This consistency shows that the shape of the c-f curve is a characteristic of dataset that can be approximated with a relatively small sample, and does not require the complete dataset to compute. The cluster number parameters used in the SparCA method for each dataset are provided in Table 1, as well as the numbers of resulting features and statistics summarizing sparsity of the resulting features.

Table 1: Classification performance, hyperparameter settings, and run time for both embedding models across tasks. For MNIST and IMDb, the performance was scored by overall accuracy in the test set, while CuMiDa was scored by average balanced accuracy across five folds of the classification set.

| Task | PCA score | PCA fit time (s) | PCA N components | SparCA score | SparCA fit time (s) | SparCA model N clusters | SparCA model N features |
|---|---|---|---|---|---|---|---|
| MNIST | 0.93 | 0.7 | 115 | 0.93 | 3.3 | 70 | 182 |
| IMDb | 0.84 | 1.4 | 184 | 0.85 | 7.8 | 175 | 340 |
| CuMiDa | 0.94 | 0.5 | 12 | 0.98 | 64.4 | 170 | 508 |

## 3.2 Supervised learning results

### 3.2.1 MNIST

Classification performance metrics on the reserved test set are summarized in Table 1. Across all metrics, the SparCA and PCA models achieve nearly identical results. This parity is to be expected as they are both based on singular value decomposition (SVD) of the data: PCA performs a single SVD of the complete dataset, while SparCA performs SVD operations separately over many disjoint subsets of the dataset features. While there is little difference between the performance of the two models for classification, the difference in interpretability of the resulting models is stark. Fig. 3 illustrates the superposition of all features weighted by their coefficients for one-vs-rest classification of the digit 8 for the SparCA and PCA based models, as well as the top 9 weighted features from each model ranked by absolute coefficient value.

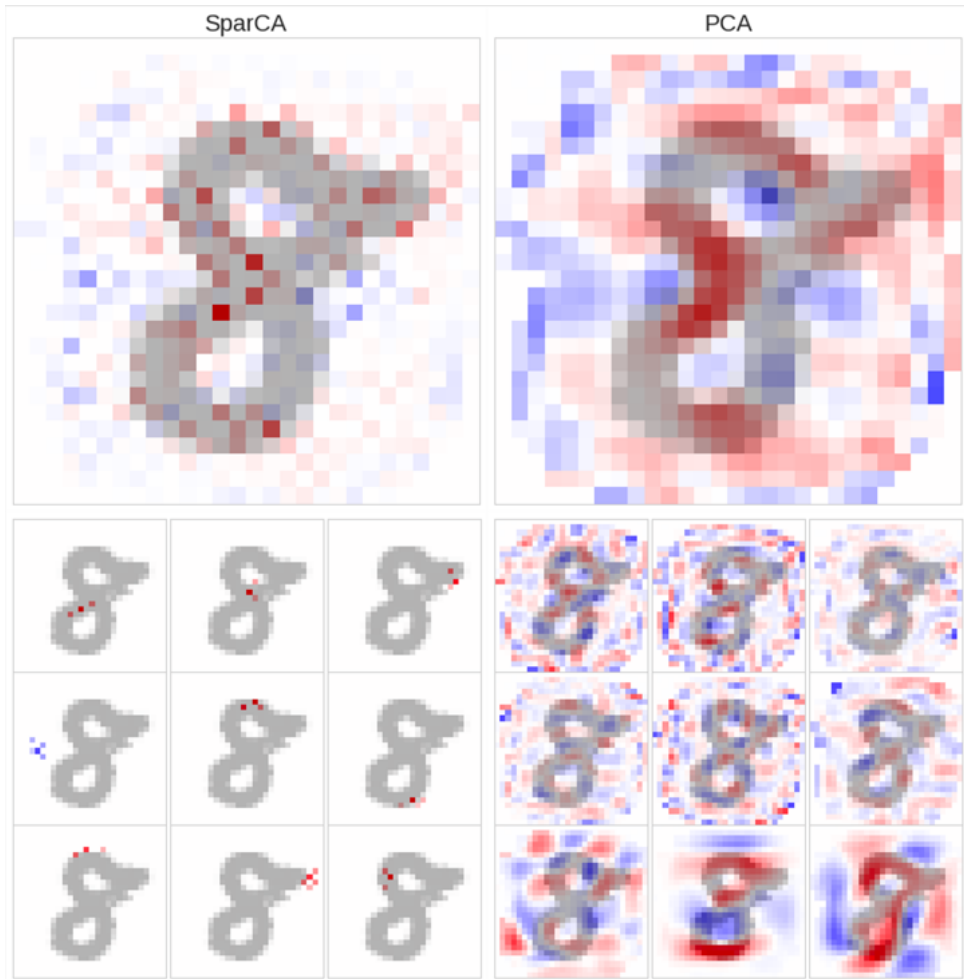

Figure 3: Overall model attention and feature-level attention for the top 9 features (sorted left to right and then top to bottom) for classifying digit 8 in the SparCA (left) and PCA (right) models.

In the case of the PCA model, each feature is a global filter that is applied over the entire image. In contrast, the SparCA features are all highly localized, which makes it quite straightforward to determine exactly what parts of the image are being used by the model to determine class membership. The combined features show that both models identify broadly similar regions of the image, but the SparCA model is much more sparse. This sparsity makes the SparCA model more robust to noise, as illustrated in Fig. 4, which shows test set accuracy under increasingly strong perturbation by Gaussian noise.

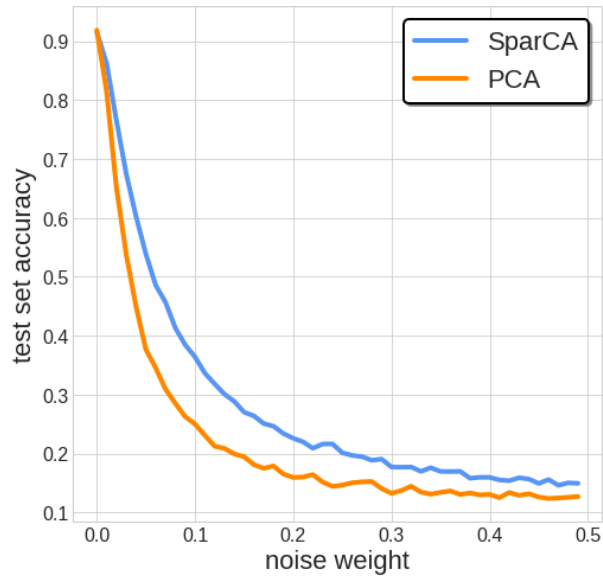

Figure 4: Model performance under perturbation by random noise.

### 3.2.2 IMDb

As with the MNIST task, the performance of the SparCA and PCA models for sentiment classification in the IMDb dataset was nearly identical, as described in Table 1, but with marked differences in interpretability. Table 2 summarizes the ten most important features for both models. For the SparCA model, all terms associated with each feature are listed. For the PCA model, because every term has a non-zero weight for each principal component, the list of terms for each feature is sorted by weight magnitude and truncated to the list of terms with weight magnitude of at least 20% of the largest absolute magnitude, up to a maximum of thirty terms.

Table 2: Top ten features for positive vs negative review classification in the IMDb dataset.

| Rank | SparCA | PCA |
|---|---|---|
| 1 | wors, stupid, terribl, aw, horribl, suck, ridicul, money, plain, save | even, get, like, thing, bad, go, kill, look, make, great, actual, tri, say, seem, guy, ani, dont, know, becaus, excel, love, whi, someth, onli, doesnt, minut, happen, look like, anyth, noth |
| 2 | wast time, wast | movi, thi movi, hi, thi, worst, watch, worst movi, movi wa, watch thi, ever seen, bad, thi movi wa, movi ever, ever, film, watch thi movi, seen, wast, young, dont, man, ha, becom, wa, wast time, perform, also, ive, play, see thi movi |
| 3 | great, fantast, amaz, brilliant, greatest, incred, favorit, alway | movi, thi movi, veri, seen, good, ever seen, ive, charact, ever, ive ever, veri good, role, thi film, play, film wa, movi ever, best, love, hi, great, job, well, thi film wa, perform, film, also, veri well, one best, stori, ive seen |
| 4 | tri, avoid, annoy, fail, unless, instead, badli, whole, mess, lame, hard, pain, stuff | year, ago, year ago, ive, seen, wast, first, music, dvd, ive seen, saw, one, thi movi, best, wast time, song, still, wa, releas, sinc, rememb, seri, fan, sing, one best, tv, im, last, movi, first time |
| 5 | best, one best | wa, movi wa, thi movi wa, horror, thi show, show, hi, wast, play, film wa, wife, role, wast time, horror film, actor, peopl, thi film wa, hi wife, career, horror movi, robert, cast, thi movi, wa made, episod, york, act wa, perform, act, year |
| 6 | bore, dull, predict, total, ok | film, thi film, effect, show, special effect, budget, act, year, direct, live, famili, low budget, thi show, low, home, special, director, kid, love, horror, script, plot, back, father, saw, old, parent, rememb, mother, boy |
| 7 | perform, cinematographi, excel, score, superb | horror, show, horror movi, thi show, horror film, year ago, ago, thi movi, saw thi, movi wa, low budget, funni, thi movi wa, budget, comedi, joke, saw, effect, gore, episod, special effect, scare, movi, year, film, charact, season, blood, scari, special |
| 8 | love thi, love | thi film, film, thi, wa, kill, read, would, horror, film wa, thi film wa, feel, think, understand, guy, peopl, murder, felt, saw, differ, see, comment, killer, thought, us, mani, view, polic, gore, get, charact |
| 9 | ever seen, ever, worst movi, worst | york, new york, year ago, ago, year, film wa, year old, thi film wa, horror, shot, citi, camera, guy, music, wife, new, bad guy, old, look like, wa, special effect, read, shoot, look, great, horror movi, special, hi wife, effect, fight |
| 10 | direct, director, script, screenplay, writer, poor | new york, york, year ago, ago, new, special effect, year, year old, saw thi, special, best, read, main charact, book, male, saw, stereotyp, adapt, novel, hi wife, know, ive, ive ever, main, budget, worst, wife, seem, effect, old |

It is evident from Table 2 that SparCA extracts concise topic areas from the corpus and summarizes them effectively with a relatively few tokens each. It is also evident that each feature represents either uniformly positive or negative sentiment, making the interpretation of each feature straightforward. In contrast, the PCA features are much denser even when truncated to the most impactful terms, and some features contains both positive and negative sentiments. Furthermore, all PCA features contain many words that are not relevant to sentiment analysis, while each term in each SparCA feature has a strong positive or negative sentiment association.

### 3.2.3 CuMiDa brain tumor cell classification

The dataset used for this task is qualitatively different from both the MNIST and IMDb review tasks in terms of size and shape: while both the MNIST dataset and IMBb datasets can be considered high dimensional, in both cases there are many more observations than features, whereas the CuMiDa microarray data consists of many more features than observations. Unsurprisingly, the SparCA algorithm provided the largest relative dimensionality reduction in this dataset, as it was by far the highest dimensional dataset of the three to begin with. In contrast with the previous two tasks, the SparCA based model performed notably better than the PCA model for this task, possibly due to the flexibility of the SparCA framework to discover more reduced features than observations, whereas PCA can extract at most $N_{samples}$ components in datasets where the number of features is larger than the number of samples.

The most important features in the SparCA based model for classifying each cell type are summarized in Table 3. For readability, only the direction of effect for each individual gene is reported, rather than the numeric gene-level weight.

Table 3: Features used to classify each tissue type in the SparCA based model. Plus (+) and minus (-) signs indicate the direction of effect for each gene in each component, while the value in the weight column provides the signed coefficient for each component.

| class | rank | weight | component genes |
|---|---|---|---|
| ependymoma | 1 | 0.79 | TEKT1(+) |
| | 2 | -0.23 | PHF11(-), PSAT1(-), PTRF(-), SLC25A5-AS1(-) |
| | 3 | 0.16 | CXorf40A(-), DACH1(+), KCNJ12(+), MDH2(-), NFRKB(+), SP5(+), ZNF276(-) |
| | 4 | 0.13 | AKAP13(+), DDX42(+), GLIS2(+), SRGAP1(+), UEVLD(+) |
| | 5 | 0.12 | 240458_at(-), C2orf74(-), C2orf88(+), CD9(+), DIRAS3(-), FAM174A(-), FAT1(+), ITGAV(-), KCND1(-), MAMLD1(+), RP11-504A18.1(-) |
| glioblastoma | 1 | 0.33 | 230959_at(+), 233295_at(-), 236908_at(+), 237246_at(+), C11orf87(-), CASC15(-), CHKB-CPT1B(+), EFNA5(-), LCE3D(-), NHLH2(-), PCDH8(-), PDE7A(+), RP1-86D1.3(+), XRCC6BP1(+) |
| | 2 | 0.32 | 1557624_at(-), CD276(+), CNOT3(-), DAXX(-), DESI2(+), EBF2(-), GLYCTK-AS1(-), NDE1(+), STK25(-), UBE3C(+), YKT6(+), ZNF354C(-), ZNF526(-) |
| | 3 | 0.23 | 91682_at(+), ALDH1A3(+), LL22NC03-N14H11.1(+), LOC441666(+), NTSR1(+), SLC13A1(+) |
| | 4 | -0.13 | 1561863_a_at(+), 233286_at(+), FAM156A(-), GTSE1-AS1(-), IGSF5(-), INTS3(+), LINC00893(+), LOC100505920(-), LOC102724814(-), NKX2-8(+), RP11-131L23.2(-), RP11-727A23.11(+) |
| | 5 | -0.13 | 235147_at(+), DIP2C(+), ELMO3(-), RNF175(+), RPF1(-), SATB1(+), SEC61B(-), SLC35G2(+), STRN(+), TGFB1I1(-), URB2(-) |
| medulloblastoma | 1 | 0.42 | 238204_at(-), BAG3(-), PPP1R14C(-), PRIMA1(-), RP11-218C14.8(-), SHISA4(-) |
| | 2 | 0.39 | FADS3(-), HEPACAM(-), LOC101927204(-), SLC18B1(-) |
| | 3 | 0.26 | 215542_at(+), BARHL1(+), CRMP1(+), LGALSL(+), MAP3K13(+), PCDHB8(+) |
| | 4 | 0.09 | EPB41L5(+), LOC101927841(+), MYB(+), PRDM13(+), RP1-86D1.3(+) |
| | 5 | 0.09 | 227571_at(+), DDX3Y(+), FAM101A(+), KCNA5(+), LOC101928605(+), PRSS27(+), VTI1A(+) |
| normal | 1 | 0.55 | 1566772_at(+), DLG3(+), GPR26(+) |
| | 2 | 0.26 | ACVR1C(+), GJB6(+), HSD11B1L(+), PLEKHG3(+), PTGDS(+), SHROOM1(+) |
| | 3 | 0.02 | AAK1(+), GRIN1(+), TMEM151B(+) |
| | 4 | 0.01 | ADAP1(+), CABP1(+), SHANK3(+), SLC45A1(+), ZDHHC23(+) |
| | 5 | 0.01 | 239199_at(+), ATP6V1G2-DDX39B(+), KAT6A(+), LOC101927268(+), RP1-193H18.2(+), SRRT(+) |
| pilocytic astrocytoma | 1 | -0.54 | 1562091_at(+), 229629_at(+), ADAMTS20(+), AIRE(-), COL4A4(-), GPR114(-), KY(-), LINC01098(+), SLC18A1(+), TRH(-), ZNF208(+) |
| | 2 | -0.23 | BC042811(-), CTD-2537I9.5(+), DBF4B(+), GMIP(-), HIST1H3B(+), MAP3K1(-), MARCKSL1(-), NLRP11(+), PTPN12(-), RP11-196G18.24(+) |

| | 3 | 0.2 | `1557348_at(+), IGSF9B(+), KLRC3(+), PCDHGA1(+)` |
| | 4 | -0.01 | `AAGAB(+), AP1S3(+), DAK(+), DNAL4(+), FAM3A(-), GEMIN4(+), LRFN4(+), MAN2B1(-), MED11(-), MEPCE(+), MMP23A(+), PEPD(-), SPNS1(-)` |
| | 5 | 0.01 | `239199_at(+), ATP6V1G2-DDX39B(+), KAT6A(+), LOC101927268(+), RP1-193H18.2(+), SRRT(+)` |

A similar table for describing the PCA based model is not practicable, as each principal component represents significant contributions of thousands of genes. In contrast, the SparCA framework produced an easily interpretable model, with features that are comprised of relatively few individual genes each.

**3.3 Run time profiling**

The time complexity characteristics of the SparCA method were measured empirically using synthetic datasets. For each measurement, synthetic data was generated to exhibit normally distributed singular values with effective rank equal to 1/5 the number of features, and the number of clusters was set to two times the effective rank. Scaling behavior with number of samples was measured using a constant value of 100 features, while a value of 50 samples was used to measure scaling with number of features. The run time results are shown in Fig. 5.

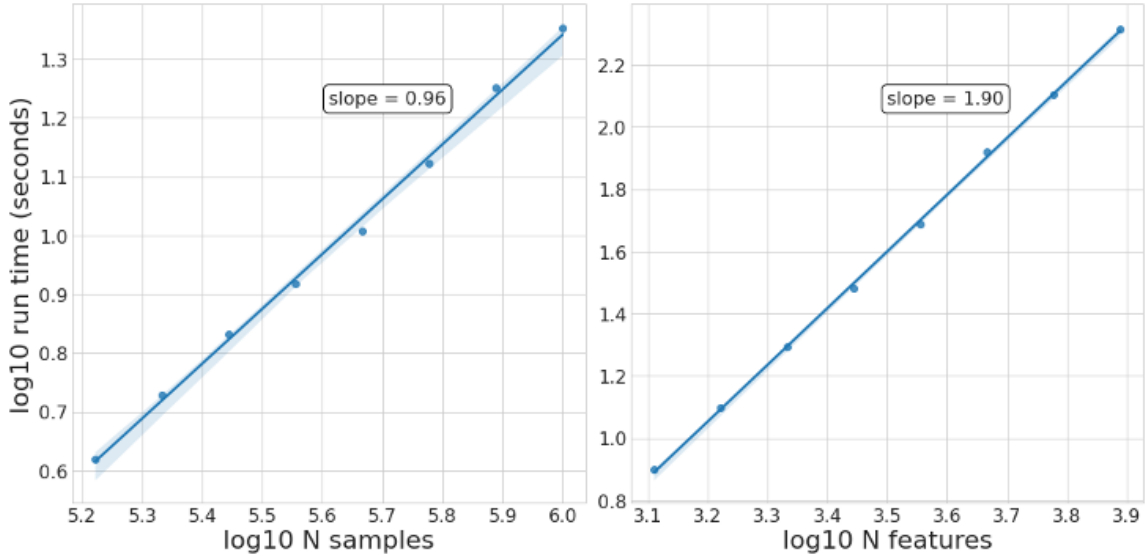

Figure 5: Run time for SparCA procedure as a function of sample number (left) and feature number (right).

In the regime where number of samples is greater than number of features, the SparCA method exhibits approximately $\mathcal{O}(N_{samples})$ behavior, and $\mathcal{O}(N_{features}^2)$ scaling where number of features is greater than number of samples.

## 4 Conclusion

In this study, we have proposed a new procedure for simultaneous dimensionality reduction and feature selection with the goal of achieving sparse, highly interpretable reduced features. Our procedure, called SparCA, achieves this through a multistep process that first agglomerates input features into a specified number of clusters that minimize within cluster variance, then finds a lower dimensional representation of each cluster by applying PCA, and finally applies OMP to reconstruct that lower dimensional representation with the fewest possible input features. The result is a linear dimensionality reduction where each feature is a sparse and compact representation of disjoint groups of input features that are individually easily interpretable. Furthermore, this is achieved without the need for hyperparameter turning against a supervised learning task, as SparCA features only two hyperparameters that are intuitively understandable and easily chosen. To demonstrate these advantages, we have benchmarked our procedure using publicly available datasets across three domains: images, natural language, and single cell gene expression. In each case, the SparCA method produced a highly interpretable feature space that equaled or exceeded the downstream classification performance of standard PCA. While PCA maintains a significant run time advantage, our experiments show that the SparCA procedure is computationally amenable to very high dimensional datasets. The SparCA method as implemented here currently has some limitations. It is not appropriate for non-continuous input features, and at its core it is a linear transformation that is similar in many respects to standard PCA. However, as a framework it might be easily extended to overcome these shortcomings, for example multiple correspondence analysis could be used for clusters composed of categorical variables, and kernel PCA could be applied to learn non-linear transformations.

**Code availability**

A python implementation of SparCA is available for use here: https://github.com/Neurology-AI-Program/sparca.git